\documentclass[11pt,a4paper]{article}
\usepackage[hyperref]{acl2019}
\usepackage{times}
\usepackage{latexsym}
\usepackage{algorithm}
\usepackage{algorithmic}
\usepackage{amsthm}
\usepackage{amsmath}
\usepackage{url}
\usepackage{amssymb}
\usepackage{graphics}
\usepackage{graphicx}
\usepackage{float}
\aclfinalcopy 

\title{Efficient text generation of user-defined topic using generative adversarial networks}

\author{Chenhan Yuan \\
  Dept. of Computer Science\\
  Virginia Tech\\
  VA, USA\\
  \texttt{chenhan@vt.edu} \\\And
  Yi-chin Huang \\
  Dept. of Computer Science \\
  National Pingtung University \\
  Pingtung, Taiwan\\
  \texttt{ychin.huang@gmail.com} \\\And
  Cheng-Hung Tsai\\
  Institute for Information Industry\\
  Taipei, Taiwan\\
  \texttt{jasontsai@iii.org.tw}}

\date{}

\begin{document}
\maketitle
\begin{abstract}
  This study focused on efficient text generation using generative adversarial networks (GAN). Assuming that the goal is to generate a paragraph of a user-defined topic and sentimental tendency, conventionally the whole network has to be re-trained to obtain new results each time when a user changes the topic. This would be time-consuming and impractical. Therefore, we propose a User-Defined GAN (UD-GAN) with two-level discriminators to solve this problem. The first discriminator aims to guide the generator to learn paragraph-level information and sentence syntactic structure, which is constructed by multiple-LSTMs. The second one copes with higher level information, such as the user-defined sentiment and topic for text generation. The cosine similarity based on TF-IDF and length penalty are adopted to determine the relevance of the topic. Then, the second discriminator is re-trained with generator if the topic or sentiment for text generation is modified. The system evaluations are conducted to compare the performance of the proposed method with other GAN-based ones. The objective results showed that the proposed method is capable of generating texts with less time than others and the generated text are related to the user-defined topic and sentiment. We will further investigate the possibility of incorporating more detailed paragraph information such as semantics into text generation to enhance the result.
\end{abstract}

\section{Introduction}
Text generation, as a basic natural language processing task, has many applications, such as dialogue robots \cite{li2017adversarial}, machine translation \cite{hu2017controllable}, paraphrasing \cite{power2005automatic} and so on. With the rise of deep learning, different neural networks are introduced to generate text. 
For example, researchers use the recurrent neural network (RNN) \cite{mikolov2010recurrent} to train the language model because of its capability to process sequential data. However, the RNN suffers from the gradient vanishing problem \cite{hochreiter1998vanishing} when the sequence becomes longer. To address this problem, Long Short-Term Memory (LSTM) \cite{hochreiter1997long}is further adopted as a sequential neural network model to generate sentences.

Lately, the Generative Adversarial Networks (GAN) framework \cite{goodfellow2014generative} has been introduced into the NLP community. GAN has two different models for completing the data-generating task. One of them is Generator G, which is responsible for generating data, and another one is discriminator D, which determines whether the input data is the real data or not. The generator G continuously optimizes generated data based on the judgment of discriminator D. After several epochs, the generated data will become more realistic.

However, GAN was originally designed to process continuous data, and using discrete data as input would make it impossible to update the gradients of the GAN framework\cite{huszar2015not}. To process discrete data, several variants of the GAN model for generating text have been proposed. These GAN variants could achieve good performances in text generation task, such as MaskGAN \cite{fedus2018maskgan}, RankGAN \cite{lin2017adversarial}, and TextGAN \cite{zhang2016generating}. 

In order to make these models fit the distribution of real text data better, the number of parameters of text generation models based on neural network are increased, which means that training these neural network models often takes a lot of time even using GPU. Conventionally, topic-related text generation models incorporate an arbitrary topic as an input by adopting mechanisms like attention \cite{feng2018topics}. Therefore, each time when the user wants to generate new sentences with another topic or sentimental tendency, the text generation models have to be retrained with all parameters to satisfy new requirements. In some scenarios, e.g., news generation, spending lots of time retraining model is not practical and the user wants new responding quickly. 

To tackle this problem, a novel text generation model based on GAN is proposed, which is called User-Defined Generative Adversarial Networks (UD-GAN). The key idea is to separate the sentence syntax model as the basic model and the topic-related model as a higher-level model, and these two could be trained independently from each other. So, when the topic or other user-defined information is modified, e.g., sentimental tendency, only one of both models needs to be retrained. In this way, once the basic syntax model is established, the following training will become much faster, since only the higher-level model has to be retrained.

In our proposed method, the discriminator is constructed based on this idea. One of the discriminators called discriminator-general, which learns to determine the proper context information and whether the input sentence is a valid syntactic structure. Another discriminator is called the discriminator-special, which ensures the output is user-defined. Inspired by SeqGAN \cite{yu2017seqgan}, we use the evaluation results of the generated text from discriminators as a reward to guide the generator to select future actions, which is to generate an updated word. 

For training the discriminator-special, it will take feature vectors as input, instead of sentences. The feature vector is defined based on the sentiment detection and topic relevance of generated sentence. The cosine similarity based on TF-IDF and length penalty are jointly adopted to represent topic relevance. 

Note that the UD-GAN is designed to be more practical to generate short paragraphs, which means sentences generated by it should be context-aware and behave like a paragraph together with surrounding sentences. To achieve this idea, discriminator-general is designed with hierarchical multiple LSTM layers. The LSTM at the top of the network processes paragraph-level information while the bottom LSTMs process sentence-level information.

The organization of the paper is as follows: First, we discussed the related works of our method in the section 2. The proposed method is described in the Section 3, including the feature extraction and model definition and training. In the Section 4, the experiment settings and evaluation results of the comparing methods are depicted. Finally, the concluding remarks and future works are described in the Section 5. 
\begin{figure*}[ht]
\centering
\includegraphics[scale=0.4]{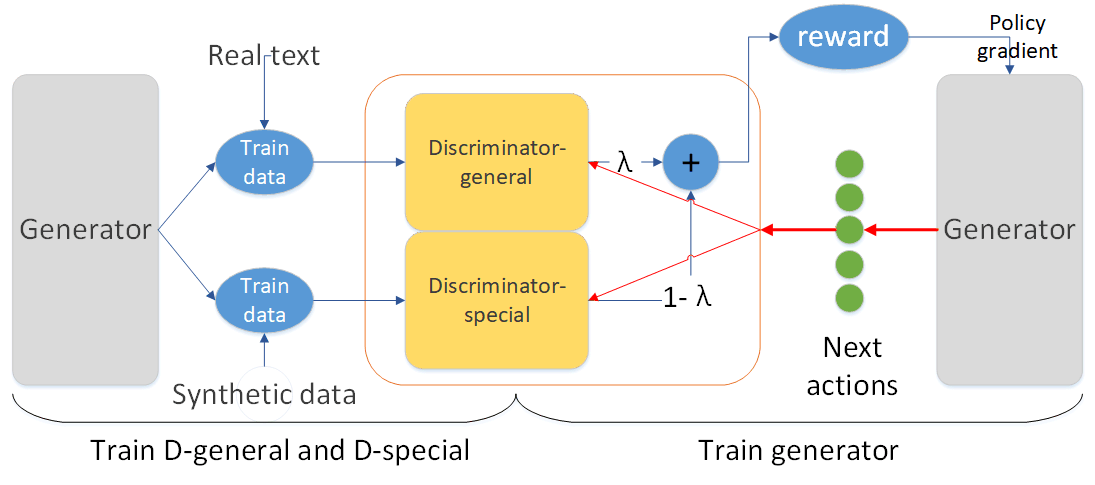}
\caption{The framework of the proposed UD-GAN}
\label{figl}
\end{figure*}
\begin{algorithm}
\caption{Initial training generator $G\theta$, discriminator-special $D\gamma$, discriminator-general $D\phi$} 
\label{alg1}
\begin{algorithmic}[1]
\STATE {Initialize $G\theta$, $D\phi$ and $D\gamma$ with random weights$\theta$, $\phi$ and $\gamma$} 
\STATE {Pre-train $G\theta$ using MLE on real text data set}
\STATE {Generate negative samples using $G\theta$ to train $D\phi$ and $D\gamma$}
\STATE {Generate synthetic positive samples to train $D\gamma$}
\STATE {Minimizing the cross entropy to pre-train $D\gamma$}
\STATE {Minimizing the cross entropy to pre-train$D\phi$}
\FOR{$i\leftarrow 1$ to $M$}
\FOR{$j\leftarrow 1$ to $N$}
\STATE {Generate a sequence $Y_{1:T}\backsim G\theta$}
\STATE {Compute rewards via Eq.\ref{Q2D}}
\STATE {Update parameters of $G\theta$ via Eq.\ref{policy_gradient}}
\ENDFOR
\FOR{$k\leftarrow 1$ to $P$}
\STATE {Generate negative samples using $G\theta$}
\STATE {Train $D\phi$ with negative samples and real text data via Eq.\ref{loss_D}}
\ENDFOR
\FOR{$l\leftarrow 1$ to $T$}
\STATE {Generate feature vectors corresponding to negative samples generated by $G\theta$}
\STATE {Generate synthetic feature vectors}
\STATE {Train $D\gamma$ with negative and synthetic feature vectors via Eq.\ref{loss_D}}
\ENDFOR
\ENDFOR
\end{algorithmic}
\end{algorithm}
\begin{algorithm}
\caption{Following training generator $G\theta$, discriminator-special $D\gamma$} 
\label{alg2}
\begin{algorithmic}[1]
\STATE {Initialize $G\theta$, $D\gamma$ with random weights$\theta$, $D\gamma$}
\STATE {Load trained $D\phi$}
\STATE {Do 2$\backsim$5 steps in Algorithm \ref{alg1}}
\FOR{$i\leftarrow 1$ to $M$}
\STATE {Do 8$\backsim$12 steps in Algorithm \ref{alg1}}
\FOR{$l\leftarrow 1$ to $T$}
\STATE {Generate feature vectors corresponding to negative samples generated by $G\theta$}
\STATE {Generate synthetic feature vectors}
\STATE {Train $D\gamma$ with negative and synthetic feature vectors via Eq.\ref{loss_D}}
\ENDFOR
\ENDFOR
\end{algorithmic}
\end{algorithm}
\section{Related Work}
Text generation is a basic task in natural language processing (NLP). In previous works, many researchers \cite{power2005automatic} extracted grammar rules from text to generate new texts. These works are capable of generating semantically rich and grammatically correct text, but due to the fixed grammar rules, generated sentences are quite lack of diversity. As neural networks could fit the distribution of real data better, some researchers design GAN-based models as language models to generate text. Unlike standard GAN, the loss function or training method of generator are modified to enable GAN to process discrete data. 

For example, In TextGAN \cite{zhang2016generating}, researchers apply feature matching with standard GAN loss function to train the generator. Reinforcement learning \cite{sutton2000policy} is another useful machine learning technique to train model with unlabeled data. Trained model will choose next actions to maximize expected reward, which is given by interface environment. Yu proposed SeqGAN \cite{yu2017seqgan}, which combine reinforcement learning with GAN. In SeqGAN, the generator uses the result of discriminator as a reward and choose next actions, which is to generate the next words in text generation task. To generate longer text, LeakGAN \cite{guo2018long} is introduced to enable the discriminator leaks features extracted from its input to generator, which then uses this signal to guide the outputs in each generation step before generating the entire sentence.

Another vital application of NLP is the sentiment analysis \cite{pang2008opinion,wilson2005recognizing}. Generally, the sentiment analysis task measures the emotional tendency of the whole sentence based on the word usage that can represent emotions in that sentence. Therefore, the establishment of an emotional word dictionary is essential. Affective Norms for English Words (ANEW) \cite{bradley1999affective} lexicon sorts all words according to rating score from 1 to 9. The highest score means the sentence convey a very positive emotion, and the lowest one represents the most negative emotion for the sentence. Based on that, some researchers \cite{hutto2014vader} construct a gold-standard list of lexical features then combine these lexical features with consideration for five general rules, which could represent the sentiment of a sentence. The VADER algorithm proposes a rule-based sentiment analyzer that has outperformed the other machine learning-based algorithms.

\section{Proposed Method}
\subsection{Basic Structure of UD-GAN}
As shown in Fig.\ref{figl}, UD-GAN contains a generator $G_\theta$ that is capable of generating context-dependent sentences and the two-level discriminators. Discriminator-general $D_\phi$ guides the generator to learn the paragraph-level information and correct syntactic structure, while discriminator-special $D_\gamma$ determines whether the generated text is related to the user-defined topic and sentiment. Discriminator-special $D_\gamma$ is trained with synthetic perfect data and generated text data, while discriminator-general $D_\phi$ is trained with real text data and generated text data. 

As we apply reinforcement learning with policy gradient to train the generator, the outputs of the two discriminators for the generated text will be combined and served as a reward to train the generator. Generator $G_\theta$ will choose the best next actions based on the reward it received. After the first training via Algorithm \ref{alg1}, the discriminator-general parameters are saved as the pre-trained model. In the subsequent trainings, we only train the parameters of the generator $G_\theta$ and discriminator-special $D_\gamma$ via Algorithm \ref{alg2}. The details about training method and structure of discriminators and generator are described as follows.

\subsection{The Framework of D-Special}
{\bf The Feature Vector of D-Special}

Discriminator-special $D_\gamma$ takes a vector containing 5 elements as input, which could represent the sentimental and topical relevance of each sentence. 

In our model, users can describe the cause and effect of an event in one sentence, which is used as the topic for generating sentences. We use the first element to represent the similarity between sentence entered by the user and generated sentence, which could also represent the user-defined topic relevance of the generated text. Based on the TF-IDF \cite{sparck1972statistical} value of each word in the sentence that the user entered and the generated sentence, the cosine similarity between these two sentences is calculated as a parameter to measure the user-defined topic relevance of the generated sentence. A larger value of cosine similarity means that the generated sentence is related to the user-defined topic.
\begin{figure*}[ht]
\centering
\includegraphics[scale=0.4]{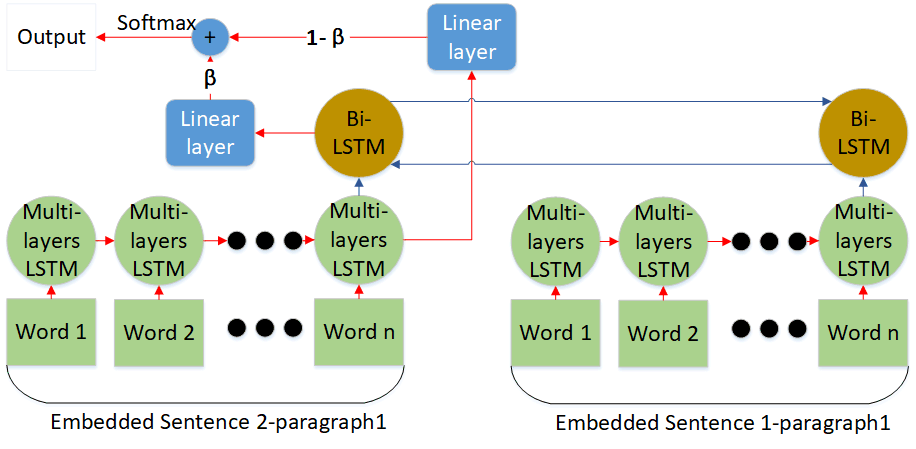}
\caption{The proposed framework for Discriminator-general}
\label{fig2}
\end{figure*}

However, if only this element is used to instruct the generator $G_\theta$ to generate topic-related sentences, the resulting sentences will be substantially as long as the user-defined topic sentence. More importantly, the generated sentences will lack diversity with same meaning. Therefore, we propose the second element, length penalty, to reduce the negative impact of the first element. The difference between the length of the generated sentence and the length of the topic sentence defined by user is mapped in [0, 1] via Eq.\ref{penalty}. 

\begin{equation}\label{penalty}
\begin{split}
& penalty_{g'} = \\
& \frac{\left| len_{g'}-len_i \right|}{ \max \limits_{g\in G} \left| len_{g}-len_i \right|-\min \limits_{g\in G} \left| len_{g}-len_i \right|}
\end{split}
\end{equation}

where i is input sentence, $g'$ is the evaluated generated sentence and G is the set of generated sentences. We set 0.5 to the optimal length penalty, which means that if the length of the sentence is very close to or very far from the length of topic sentence, it is unqualified.

We implemented the VADER algorithm to calculate the probability that a generated sentence belongs to a positive, negative or neutral emotion class. As VADER gives three values that correspond to the probability of each sentiment category, the sum of which is 1, these three values will be saved in the third to fifth elements. The optimal sentiment is defined by the user.

In conventional GAN training, the discriminator treats real text data as the positive sample and generated text as the negative sample. However, there is no sentence in real corpus that has  exactly the same features as the positive sample, since its feature vector is constructed by applying the above mention algorithm, while the user-defined feature vector is a specific value. Therefore, we train the discriminator-special $D_\gamma$ with synthetic data, which is treated as positive sample. For example, supposing that the user would like to generate an essay with one positive emotion, then the UD-GAN will generate [1,0.5,1,0,0] vectors corresponding to the number of generated sentences, which will be combined with vectors corresponding to the generated sentences as the input of discriminator-special.

{\bf The Structure of D-Special}

Two linear layers with Relu as the activation function are used as discriminator-special $D_\gamma$. The output of this network will be part of the reward to train generator $G_\theta$ after it passed through a softmax layer. 

We explain here why the multiple fully connected layer is implemented as a discriminator-special. The first reason is that after the Discriminator-General is constructed, in the subsequent training, the discriminator-special will be continuously retrained when demands of user change. This requires spending as little time as possible to train a good discriminator-special. The multiple fully connected layer has fewer parameters, which means this network will converge faster than others will. Another reason is that the aim of training discriminator-special is to distinguish whether the input vector corresponds the user-defined one. For an input with only five variables, a neural network with two fully connected layers is complicated enough to determine the class of input vector correctly.
\subsection{The Framework of D-General}
Unlike conventional ideas of using classifier-based models as a discriminator, the discriminator-general $D_\phi$ needs to process sequence data and context information, such as the paragraph information for each sentence to generate paragraph-level text.

Therefore, as shown in Fig.\ref{fig2}, we designed a hierarchical-multiple-LSTM neural network as the discriminator-general $D_\phi$. The bottom multi-layers LSTM takes an embedding vector for each word in a sentence as the input and it outputs a feature matrix representing the corresponding sentence. The top bidirectional LSTM \cite{graves2005framewise} takes the feature matrices of these sentences, which belong to the same paragraph, as input and it  outputs a feature matrix representing that paragraph. After transforming through two different linear layers respectively, the above two feature matrices will be combined together. Finally, the discriminator-general calculates the score of the input sentence via Eq. \ref{result_D}.

\begin{equation}\label{result_D}
R(Y)=softmax[(1-\beta)LSTM_\alpha+\beta LSTM_\eta]
\end{equation}
where $\beta$ is a trainable parameter ranging 0-1.
\subsection{Generator}
Generator $G_\theta$ is designed with GRU \cite{chung2014empirical}. In UD-GAN, due to the excessive parameters of the two discriminators, it is easy to guide the generator to be over-fitting. As a commonly used variant of LSTM, GRU avoids this over-fitting problem. In addition, having fewer parameters than conventional LSTM allows GRU to take less time to converge, which is the first priority in UD-GAN.
\subsection{Reward and Policy Gradient Training}
The reinforcement learning has been incorporated to enable GAN to process discrete data. In this scenario, generator $G_\theta$ will use the results from discriminators on the generated text as reward to generate next words. In UD-GAN, the reward is calculated based on results of two discriminators. Generator $G_\theta$ tries to maximize expected reward from the initial state till the end state via Eq.\ref{loss_func_generator}(loss function).
\begin{equation}\label{loss_func_generator}
\begin{split}
 &J(\theta)= \sum_{t=1}^{T}\mathbb{E}(R_t|S_{t-1},\theta) \\
 &= \sum_{t=1}^{T}G_\theta(y_t|Y)[\lambda(D_\phi(Y))+(1-\lambda)D_\gamma(Y))] 
\end{split}
\end{equation}

Where $\lambda$ is a manually set weight and Y is a complete sequence and $R_t$ is the reward for a whole sequence. In our experiments, we set $\lambda$ to 0.8 to give more weight to the discriminator-general $D_\phi$ for generating sentences with better syntactic structure. Note that since discriminators can only make the judgement with a complete sequence, the Monte Carlo search \cite{silver2016mastering} is adopted to find out some of the possible generated complete sequences of each state. the average judgment results of the discriminators for these sequences are calculated as a reward of this state.

In this paper, we implemented policy gradient method. The gradient of Eq.\ref{loss_func_generator} can be derived approximately as follows:
\begin{equation}\label{policy_gradient}
\begin{split}
&\nabla_\theta J(\theta)\simeq \\
&\sum_{t=1}^{T}\mathbb{E}_{y_t\sim G_\theta}[\nabla_\theta logG_\theta(y_t|Y)Q_{D_\phi ,D_\gamma}^{G_\theta}(y_t|Y)]
\end{split}
\end{equation}

where $Q_{D_\phi ,D_\gamma}^{G_\theta}(y_t|y_{1:t-1})$ can be derived via Eq.\ref{Q2D}.
\begin{equation}\label{Q2D}
Q_{D_\phi ,D_\gamma}^{G_\theta}(y_t|y_{1:t-1})=\lambda(D_\phi(Y))+(1-\lambda)D_\gamma(Y))
\end{equation}
The loss function of both discriminators is introduced as follows:
\begin{equation}\label{loss_D}
J=-(\mathbb{E}_{Y\backsim P_{data}}[R(Y)]-\mathbb{E}_{Y\backsim G_\theta}[1-R(Y)])
\end{equation}
where $R(Y)$ is the reward from two discriminators for a whole sequence.

\section{Experimental Analysis}
\subsection{Dataset}
We crawled nearly 10,000 press released from the opinion section of Newsweek as the training corpus. The opinion section of Newsweek is selected as training corpus because the paragraphs of the essays in Newsweek are generally closely related and not long. The other reason is that through the articles in the opinion section, authors can often convey their own sentiment tendencies. 

NER is used to replace name-entities with their name-entity tags to decrease vocabulary. After tokenizing the corpus, long sentences of more than 45 words in the corpus were removed. The final training corpus has 425K sentences and 103K paragraphs.

\subsection{Experimental Setting}
SeqGAN and LeakGAN are used as the baseline system to evaluate UD-GAN. We train SeqGAN and LeakGAN for 20 epochs, which is same as the number of times UD-GAN is trained. Other parameters of baselines remain unchanged as implemented in their original papers. 

The bottom of the discriminator-general consists of three layers of LSTM. The hidden dimension of discriminator-general bidirectional LSTMs for the UD-GAN and the bottom LSTMs is set to 64. Besides, the hidden dimension of discriminator-special linear layer and GRU unit of generator is set to 32. In each epoch of initial training, generator G is trained once, and the  discriminator-general is trained four times while the discriminator-special is trained twice. 

For evaluating the effectiveness of the proposed method, we first compared the sentences relevance to user-defined topic and sentimental tendency, and then compare the training time of each system. Finally, the fluency and correctness of UD-GAN and baseliens were evaluated.
\subsection{Relevance of Topic and Sentiment}
{\bf Relevance of Topic}

As an objective summary accuracy evaluation method that is widely used, ROUGE \cite{lin2004rouge} is also adopted here to evaluate whether generated sentences are related to user-defined topics.
\begin{table}
\begin{center}
\begin{tabular}{lrr}
\hline
\bf GAN-based models & \bf ROUGE-L\\
\hline
UD-GAN(GS) & 364.73  \\
UD-GAN(S) & \bf 370.54 \\
UD-GAN(G) & 340.19 \\
SeqGAN & 342.27 \\
LeakGAN & 345.03 \\
\hline 
\end{tabular}
\caption{The ROUGE-L score for each system. UD-GAN(G+S) represents initial training and UD-GAN(S) represents following training. UD-GAN(G) only has discriminator-general and generator. Note that this score is the sum of all generated sentences' ROUGE-L results.}
\label{ROUGE}
\end{center}
\end{table}
Generated sentences are treated as summaries to be evaluated, and the topic sentence defined by user is used as a reference summary to evaluate whether the generated sentence is related to the topic. Note that even if the ROUGE scores of the generated sentences are not high, it does not mean that these sentences are not closely related to the user-defined topic necessarily. One possibility is that the generated sentences will use other words or syntactic structures to describe the topic sentence. 

In this paper, we report the sum of ROUGE-L scores of all sentences. Based on the longest common subsequence, ROUGE-L is a score related to recall rate. As shown in Table.\ref{ROUGE}, the ROUGE-L scores for UD-GAN (G+S) and UD-GAN(S) are slightly higher than baseline systems and UD-GAN (G).

{\bf Relevance of Sentimental Tendency}

The VADER algorithm is used to calculate the probability that the sentimental tendency of the generated sentences to be positive, negative or neutral. Here, we evaluated the system performance by setting the target sentimental tendency as positive. 

As shown in Table.\ref{emotion}, the average probability in each sentimental tendency category of all sentences is calculated. With training discriminator-special, UD-GAN (G+S) and UD-GAN (S) are more likely to generate positive sentences than baselines. Which proves that the proposed method is capable to generate the sentences with the desired sentiment. However, since the total number of sentences expressing positive sentimental tendency in the training corpus is quite low, the probability of UD-GAN generating positive sentiment is still not higher than 0.5.
\begin{table}
\begin{center}
\begin{tabular}{llll}
\hline
\bf & \bf Positive & \bf Negative & \bf Neutral\\
\hline
UD-GAN(GS) & 0.39 & 0.05 & 0.56  \\
UD-GAN(S) & \bf 0.41 & \bf 0.04 & \bf 0.55 \\
UD-GAN(G) & 0.10 & 0.08 & 0.82 \\
SeqGAN & 0.09 & 0.08 & 0.83 \\
LeakGAN & 0.08 & 0.07 & 0.85 \\
\hline 
\end{tabular}
\caption{ The probability of sentiment tendency of generated sentences}
\label{emotion}
\end{center}
\end{table}

{\bf Generate Context-dependent Sentences}

To demonstrate that UD-GAN can generate context-dependent sentences, we show sentences generated by UD-GAN and baselines. As shown in Table \ref{demo}, one can see that the proposed UD-GAN does generate sentences related to the user-defined topic. UD-GAN tries to add some conjunctions when generating sentences so that the sentences seem to be related, and each sentence is extended with other related words based on the topic. Note that there are some Name-Entity (NE) tags generated by the models because the NE tagging has been done for simplifying the corpus lexicon.

However, semantically, these sentences are not intrinsically related to each other, which is a problem we will address in the future.
\begin{table}

\begin{tabular}{llll}
\hline
\bf {topic:} the attack in douma occurred days\\
\bf after trump indicated that he wanted to pull\\
\bf us troops out\\
\hline
\bf UD-GAN(S): \\
1. the country contacts to the u.s. and trains  \\
{\bf troops} for government living on the federal \\
system in LOCATION . \\
2. we are discussed actively : if u.s. is the facts \\
that citizens in the country will likely vote \\
for type elections ? \\
3. during these {\bf attack} things {\bf occurred days} , i \\
say just PERSON who {\bf pulls} in the exchange \\
best {\bf troops out} as trade in LOCATION . \\
4. and he often enthusiastic , telling only having\\
heard nothing happened while you can indicate \\
to {\bf pull out} from country . \\
5. but these generations in LOCATION can\\
predict the next five {\bf attacks occur}.\\
\hline 
\bf LeakGAN: \\
1. it prompted the opposition during a `` real ''\\
of subtlety , and video straws . \\
2. but if PERSON know that we serve the best \\
drives these country purposes . '' \\
3. besides disarming our administration and\\
pricing and its traditional views . \\
4. with her contempt for all enough neighbors . '' \\
5. one day i 'd go beyond my candor .\\
\hline
\bf SeqGAN: \\
1. we do n't mean . \\
2. you should be `` changed '' that you know .\\
3. i 've always been proposing the findings .\\
4. in other words , he 's because you have a \\
testament to his goodness -- not a result .\\
5. he gave economic law .\\
\hline 
\end{tabular}
\caption{An example of the generated sentences from different systems}
\label{demo}

\end{table}
\begin{table}
\begin{center}
\begin{tabular}{lrr}
\hline
\bf GAN-based models & \bf Time s\\
\hline
UD-GAN(GS) & 29061.48 \\
UD-GAN(S) & \bf 4841.99 \\
UD-GAN(G) & 29036.65 \\
SeqGAN & 27011.08 \\
LeakGAN & 30471.95 \\
\hline 
\end{tabular}
\caption{Time spending on training of each models}
\label{time}
\end{center}
\end{table}
\subsection{Training Time Evaluation}
The time spending on gradient propagation and update of UD-GAN and baselines are compared, instead of the time spending on loading and saving data. Our platform is a workstation with a GeForce GTX 1080 Ti graphics card with 11G RAM. All GAN-based models compared here are implemented in pytorch \cite{paszke2017automatic} framework to eliminate the impact of different frameworks on time consumption. 

As shown in Table.\ref{time}, because the structure of discriminator-general is more complex than the structure of discriminator D of baselines, initial training of UD-GAN takes the longest time. However, in the subsquent trainings, due to the gradient propagation and parameter update of discriminator-special is quite fast, the time required to train UD-GAN (S) is the shortest. The UD-GAN (S) takes only about an hour and a half to complete training, which is much less than the nearly eight hours of training time for baselines.

\subsection{Fluency and Accuracy}
As shown in table \ref{BLEU}, we report BLEU \cite{papineni2002bleu} scores of UD-GAN and baselines to compare the fluency and accuracy of text they generate. The BLEU we use here is the average value of 1-gram BLEU, 2-gram BLEU and 3-gram BLEU, which are given the same weights . 

In the case of training the discriminator-general only, the BLEU score of the UD-GAN (G) is between SeqGAN and LeakGAN. Therefore, the accuracy and fluency evaluation of using multi-layer LSTMs as a discriminator is comparable to that of using a classifier-based model, such as CNN, as the discriminator. When the  discriminator-general and discriminator-special are simultaneously trained (initial training), UD-GAN (G+S) has a slightly higher BLEU score than UD-GAN (G). That is to say, even if discriminator-special is added and the result of discriminator-general, which can distinguish the correctness of the sentence, is less weighted, the resultant generator of UD-GAN (G+S) can still learn how to generate a sentence with the correct syntax. Then we change the user-defined topic and sentimental tendency to train the discriminator-special only (subsequent  training). The results showed that the BLEU score of the UD-GAN(S) is still between LeakGAN and SeqGAN. It means that retraining the discriminator-special has no effect on whether the generator can learn the correct syntax without changing the weights of rewards generated by discriminator-general and discriminator-special.

\begin{table}
\begin{center}
\begin{tabular}{lrr}
\hline
\bf GAN-based models & \bf BLEU score\\
\hline
UD-GAN(G+S) & 0.6412  \\
UD-GAN(S) & 0.6409 \\
UD-GAN(G) & 0.6357 \\
SeqGAN & 0.6303 \\
LeakGAN & \bf 0.7161 \\
\hline 
\end{tabular}
\caption{The average BLEU score for each system. Note that UD-GAN(S) achieves comparable BLEU performance with baselines, whose training needs far less time than baselines.}
\label{BLEU}
\end{center}
\end{table}

\section{Conclusion and Future Work}

In this paper, we propose a UD-GAN method to re-train text generation model more efficiently to generate sentences that are consistent with the new user-defined topic and sentimental tendency. We compared the accuracy and fluency of sentences generated by UD-GAN with other GAN-based text generation models. The experimental results showed that sentences generated by UD-GAN are competent. Meanwhile, UD-GAN takes much less time in the re-train stage than other models. According to experimental results, UD-GAN can also successfully generate sentences related to the user-defined topic and sentimental tendency, while baselines does not have this capability. Besides, UD-GAN can also generate paragraph-level text. 

However, the sentences generated by UD-GAN are still inferior to the state-of-the-art method, i.e., LeakGAN, in terms of fluency. And the current paragraph-level information used here does not include complex linguistic information, such as the order of sentences. In future work, we will try to maintain the existing advantages of UD-GAN while improving the readability of generated text. 

\bibliography{acl2019}
\bibliographystyle{acl_natbib}

\end{document}